\documentclass[runningheads]{llncs}
\usepackage{graphicx}
\usepackage{hyperref}
\hypersetup{colorlinks=true,citecolor=teal}
\usepackage{xcolor}
\usepackage{tikz}
\usepackage{comment}
\usepackage{diagbox}
\usepackage{booktabs}
\usepackage{amsmath,amssymb} 
\usepackage{bm}
\usepackage{multirow} 
\usepackage{subcaption}
\usepackage{color}

\usepackage[accsupp]{axessibility}  

\def\eg{\emph{e.g.}} 

\def\ie{\emph{i.e.}}

\def\etal{\emph{et al.}}

\makeatletter
\def\@fnsymbol#1{\ensuremath{\ifcase#1\or \dagger\or \ddagger\or
\mathsection\or \mathparagraph\or \|\or **\or \dagger\dagger
\or \ddagger\ddagger \else\@ctrerr\fi}}
\makeatother

\begin{document}
\pagestyle{headings}
\mainmatter
\def\ECCVSubNumber{2645}  

\title{End-to-end Graph-constrained Vectorized Floorplan Generation with Panoptic Refinement} 


\titlerunning{Vectorized
Floorplan Generation with Panoptic Refinement}
%
\author{Jiachen Liu\inst{1}\thanks{These authors contributed equally to this work.}\index{Liu, Jiachen} \and
Yuan Xue\inst{2}$^\dagger$\index{Xue, Yuan}
\and
Jose Duarte\inst{3}\index{Duarte, Jose}
\and
Krishnendra Shekhawat\inst{4}\index{Shekhawat, Krishnendra}
\and \\
Zihan Zhou\inst{5}\index{Zhou, Zihan}
\and
Xiaolei Huang\inst{1}\index{Huang, Xiaolei}}

\authorrunning{J. Liu et al.}
%
\institute{College of Information Sciences and Technology, The Pennsylvania State University
\and
Department of Electrical and Computer Engineering, Johns Hopkins University
\and
College of Arts and Architecture, The Pennsylvania State University
\and
Department of Mathematics, BITS Pilani
\and
Manycore Tech Inc.}
\maketitle

\begin{abstract}
The automatic generation of floorplans given user inputs has great potential in architectural design and has recently been explored in the computer vision community. However, the majority of existing methods synthesize floorplans in the format of rasterized images, which are difficult to edit or customize. In this paper, we aim to synthesize floorplans as sequences of 1-D vectors, which eases user interaction and design customization. To generate high fidelity vectorized floorplans, we propose a novel two-stage framework, including a \textit{draft stage} and a multi-round \textit{refining stage}. In the first stage, we encode the room connectivity graph input by users with a graph convolutional network~(GCN), then apply an autoregressive transformer network to generate an initial floorplan sequence. To polish the initial design and generate more visually appealing floorplans, we further propose a novel panoptic refinement network (PRN) composed of a GCN and a transformer network. The PRN takes the initial generated sequence as input and refines the floorplan design while encouraging the correct room connectivity with our proposed geometric loss. We have conducted extensive experiments on a real-world floorplan dataset, and the results 
show that our method achieves state-of-the-art performance under different settings and evaluation metrics.
\end{abstract}


\section{Introduction}

House design is essential yet challenging work for professional architects, usually requiring extensive collaboration and multi-round refinement. Floorplan design is a crucial part of house design, which involves designing the room layouts and their connectivities such as walls and doors. With the availability of several large-scale floorplan benchmarks~\cite{cruz2021zillow,fan2021floorplancad,wu2019data} and advance in generative models such as generative adversarial networks~(GANs)~\cite{goodfellow2014generative}, generating floorplans automatically has recently attracted the attention and interest of architects as well as computer vision researchers~\cite{huang2018architectural,newton2019generative}.
\begin{figure}
\begin{subfigure}[t]{0.34\linewidth}
\centering
\includegraphics[width=0.70\linewidth]{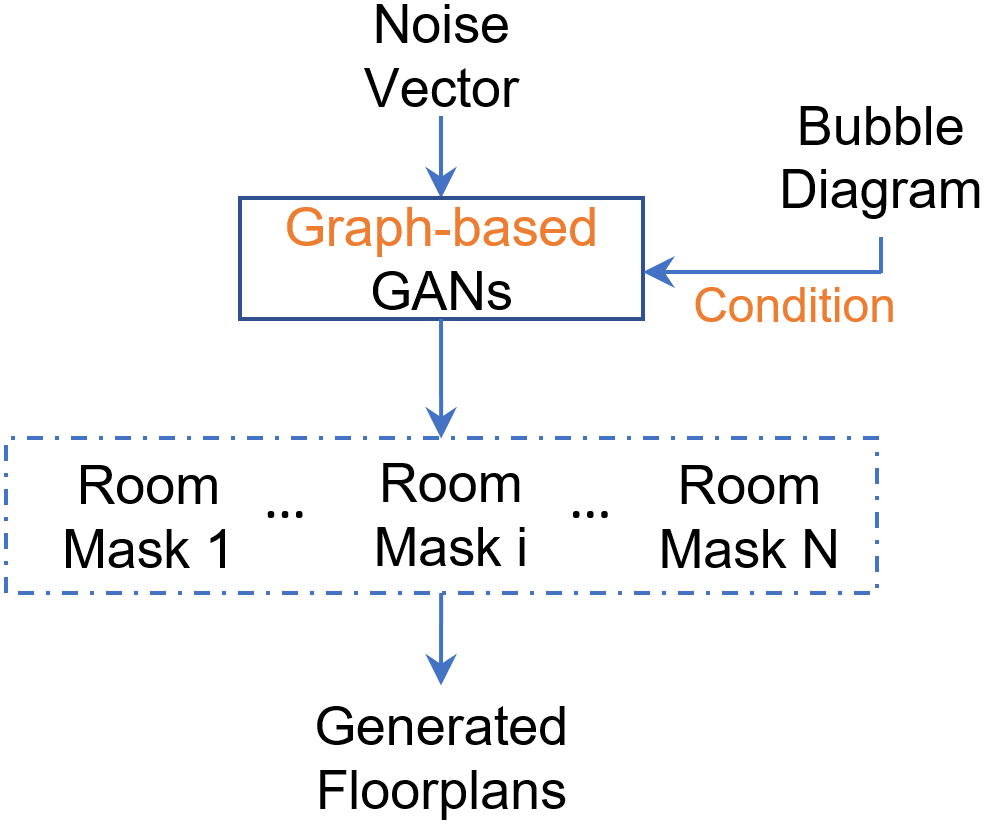}
\subcaption{Previous rasterized methods~\cite{nauata2020house,nauata2021house}.}
\end{subfigure}
\hfill
\begin{subfigure}[t]{0.28\linewidth}
\centering
\includegraphics[width=0.60\linewidth]{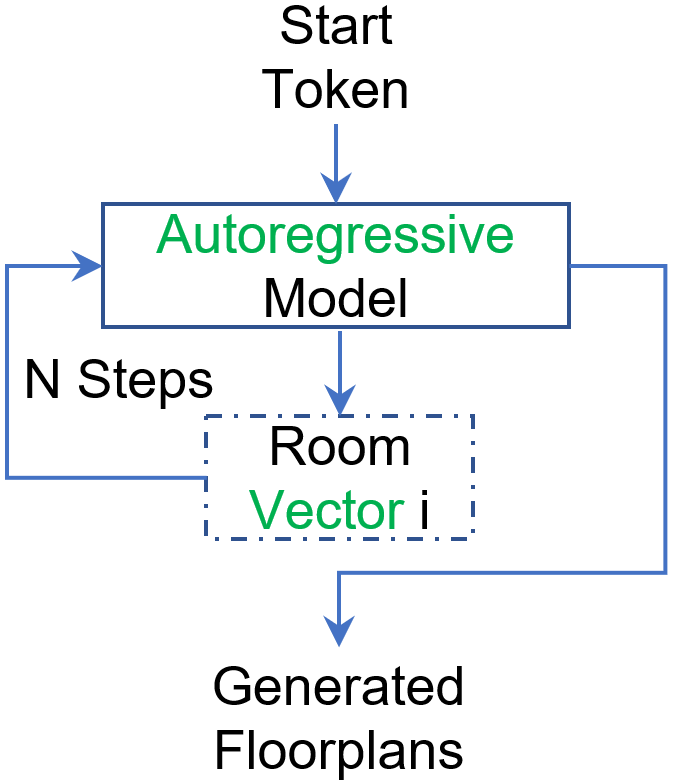}
\subcaption{Previous vectorized methods~\cite{para2021generative}.}
\end{subfigure}
\hfill
\begin{subfigure}[t]{0.36\linewidth}
\centering
\includegraphics[width=0.58\linewidth]{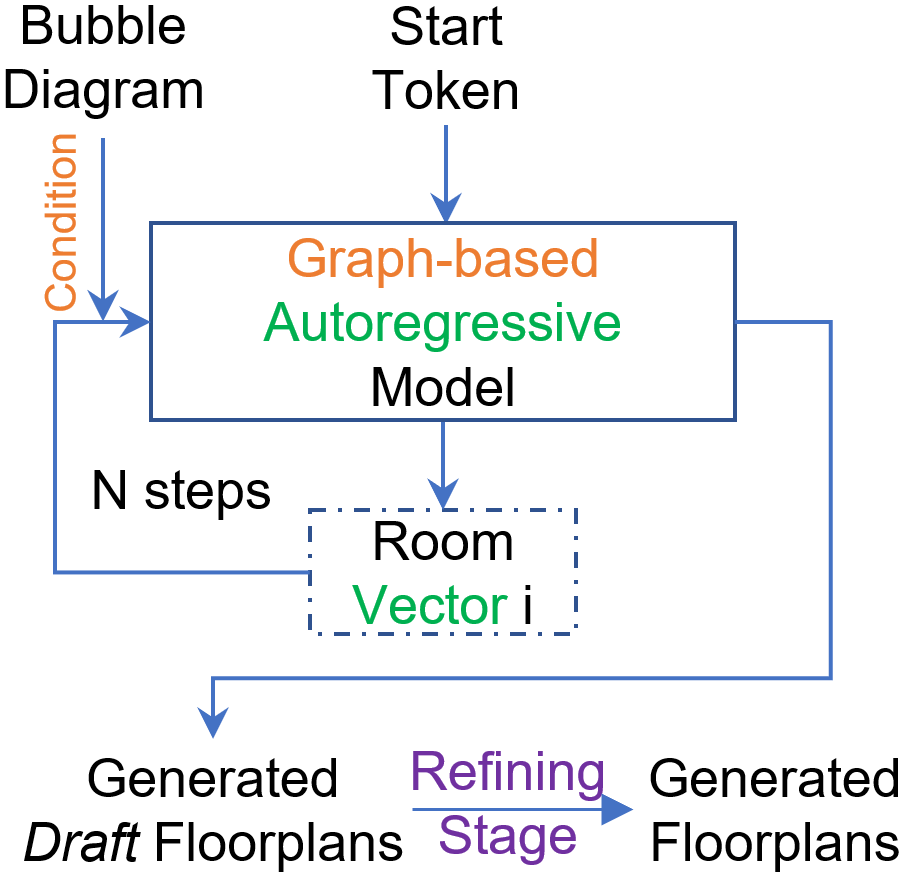}
\subcaption{Our proposed vectorized method.}
\end{subfigure}
\caption{Comparisons between different floorplan generation pipelines.}\label{fig:teaser}
\end{figure}

Motivated by the interactive design process, previous works generated floorplans with different input constraints, such as building boundaries~\cite{hu2020graph2plan,wu2019data}, room dimensions~\cite{shekhawat2021tool}, or bubble diagrams that describe room numbers, room types, and adjacency or connectivity information~\cite{nauata2020house,nauata2021house}. There is also work which explores floorplan generation from scratch in an unconditional way (\eg{}, \cite{para2021generative}).
Since architects usually preset room categories and their connectivities before designing the floorplans, we find the setting by HouseGAN~\cite{nauata2020house} to be the closest to the design practice. We thus follow their setup throughout this paper for interactive floorplan generation.
\par
Current state-of-the-art methods with conditional inputs~\cite{nauata2020house,nauata2021house} use graph-based GANs to generate room layouts as rasterized images. Although they have shown superior generation realism, we argue that GAN-generated rasterized images have several limitations in representing floorplan layouts. First, it is challenging for GANs to learn geometric layout properties such as axis-aligned walls. Second, some input conditions such as room numbers cannot be explicitly expressed in the rasterized floorplans. Last but not least, rasterized masks are irregular representations, which have restricted the options for users or architects to refine or customize the generated floorplans. In this paper, we propose to represent floorplan layouts as sequences of 1-D vectors describing the room bounding box coordinates, and generate vectorized floorplans in an end-to-end fashion. With a concise and user-friendly representation, generating vectorized floorplans enables efficient inference, more accessible user interactions, and connected space representation.
\par
Mimicking the floorplan design process, we propose a novel two-stage framework conditioned on input bubble diagrams and gradually refine the generated floorplan through multiple steps. Fig.~\ref{fig:teaser} illustrates the difference between our proposed framework and previous methods. In the first stage~(\ie{}, \textit{draft stage}), we apply graph convolutional networks~(GCNs) to encode the room connectivity from the bubble diagram, then an initial room sequence is generated with an autoregressive self-attention transformer network following previous layout generation works~\cite{gupta2021layouttransformer,para2021generative}. However, the generated draft result may be unsatisfactory since the autoregressive models cannot access the global information from the entire sequence at each generation timestamp. Thus, in the second stage (\ie{}, refining stage in Fig.~\ref{fig:pipeline}), we propose a panoptic refinement network~(PRN) to refine the draft floorplan design for a more visually appealing appearance while keeping the correct room connectivity. Taking the entire floorplan sequence from the draft stage as input, a GCN in the panoptic refinement network first processes the generated sequence to refine the connectivity relationship. Another transformer is further applied to exploit the contextual information of the whole sequence. In addition, a novel geometric loss based on layout overlapping is proposed to encourage the correct layout of the generated floorplan.
\par
The main contributions of our paper can be summarized as follows:
\begin{itemize}
\item We propose a novel two-stage framework to generate vectorized floorplans from user inputs. To the best of our knowledge, our work is the first to allow the end-to-end conditional generation of vectorized floorplans.
\item In the draft stage, we integrate GCNs into a transformer-based autoregressive model and show how to effectively capture the room connectivity relationships provided in the user inputs.
\item In the refining stage, we propose a panoptic refinement network for multi-round refinements of the draft floorplans, while encouraging the correct room connectivity with our proposed geometric loss.
\item Extensive experiments on a real-world floorplan dataset show that our method outperforms previous state-of-the-art approaches under most settings.
\end{itemize}

\section{Related Work}

\subsubsection{Data-driven floorplan generation}
Various data-driven approaches have been proposed in recent years for automatic floorplan generation. Mainstream methods primarily focused on generating floorplans as rasterized images. RPLAN~\cite{wu2019data} created a large-scale annotated dataset from residential buildings and proposed a two-stage approach to generate rooms from a given boundary. Graph2Plan~\cite{hu2020graph2plan} proposed to first generate layout-constrained floorplans as rasterized images via convolutional neural networks~(CNNs) and graph neural networks~(GNNs), then applied an offline optimization algorithm to align the rooms and transform them into a vectorized floorplan, given the boundary and graph-constrained layout as input. HouseGAN~\cite{nauata2020house} took the bubble diagram of the rooms as user input, and introduced a GAN-based architecture to generate each room. HouseGAN++~\cite{nauata2021house} was the extension of HouseGAN and the authors proposed an improved GAN-based network, using a GT-conditioning training strategy as well as a list of test-time meta-optimization algorithms.
\par
While previous rasterized floorplan generation methods have achieved promising results, their applicability is often limited by the rasterized representations. In practice, architects mostly work with vectorized representation~\cite{xu2021blockplanner} of floorplans due to its flexibility and geometric compatibility. To leverage vectorized representations, existing methods~\cite{nauata2021house,wu2019data} often converted rasterized floorplans into vectorized representations via postprocessing steps~\cite{chen2019floor}. However, the representation power is yet limited by the originally generated rasterized images. To this end, we directly represent the floorplan rooms as vectorized bounding boxes in our method. More recently, Para~\etal{}~\cite{para2021generative} also utilized vectorized representation to generate floorplans using an autoregressive model and focused more on unconditional generation. More specifically, a transformer was used to predict elements including room coordinates and sizes in an autoregressive manner, followed by another transformer to generate edges such as walls and doors. A post-processing optimization step was adopted to refine the output from the networks to obtain final layouts. Despite using the same floorplan representations, our work has several essential differences from~\cite{para2021generative}. First, our model enables interaction with graph-based user input, which satisfies the room connectivity constraints. Second, we encode constraints such as doors into the learning objectives of the same network, skipping the need to use a separate model to learn constraints as in~\cite{para2021generative}. Moreover, unlike~\cite{para2021generative} which applied linear programming regularization to optimize network output, we propose a panoptic refinement network that refines the draft floorplan while encouraging the correct room connectivity in an end-to-end fashion.
\subsubsection{Autoregressive models for layout generation}
Floorplan generation can be interpreted as a special case of layout generation. In the literature, the autoregressive model has been widely used in layout or shape generation tasks \cite{arroyo2021variational,gupta2021layouttransformer,jyothi2019layoutvae,kong2021blt,nash2020polygen} due to its superior representation ability to model sequential relationships. LayoutVAE~\cite{jyothi2019layoutvae} proposed a variational auto-encoder~(VAE)~\cite{kingma2013auto} architecture conditioned on priors such as element numbers and types. LayoutTransformer~\cite{gupta2021layouttransformer} applied a transformer-based architecture to learn layout attributes autoregressively in an unconditional manner. VTN~\cite{arroyo2021variational} instantiated the VAE framework with a transformer network and generated layouts in a self-supervised manner. However, there are two potential limitations for transformer-based autoregressive models in layout generation~\cite{kong2021blt}. First, the information flow is unidirectional and immutable in the autoregressive model since the model only has access to previously generated results during the iterative generation. Second, most existing transformers only considered unconditional layout generation without user inputs and thus are not suitable for an interactive generation. To mitigate such limitations, we enable interactive generation by integrating GCNs into the transformer for draft floorplan generation. Unlike existing refinement method for scene layout generation which only refines the output once~\cite{yang2021layouttransformer}, we resolve the unidirectional generation issue by applying another panoptic refinement network that utilizes the whole vectorized floorplan sequence and perform multi-round refinement for the draft floorplan.

\section{Methodology}
\begin{figure*}[!tp]
\centering
\includegraphics[width=0.95\linewidth]{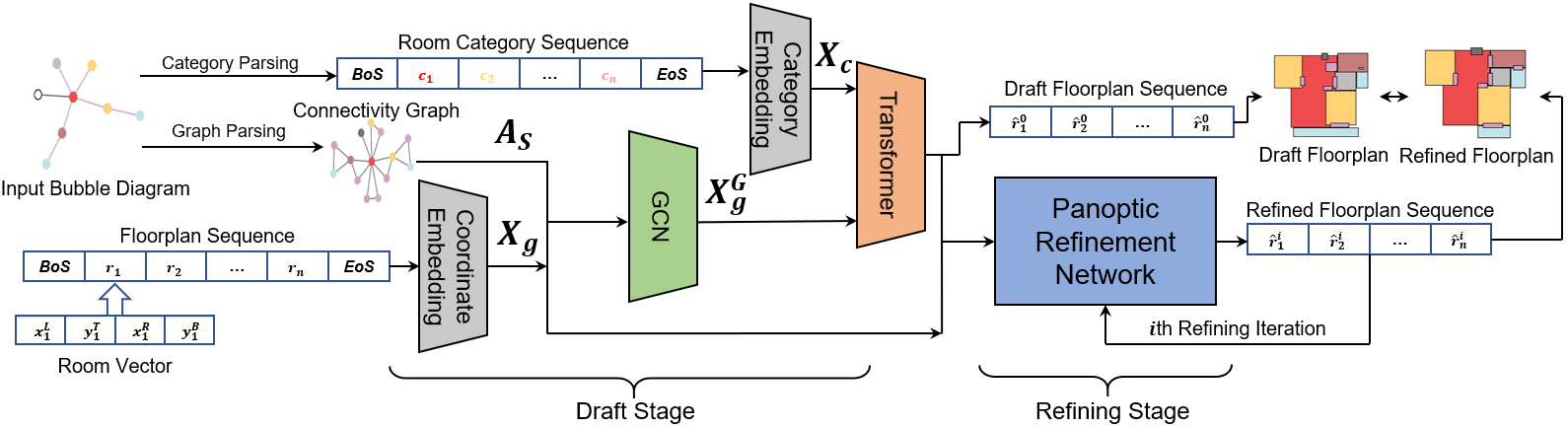}
\caption{Overview of our method. User inputs are first processed by a GCN and room elements are generated by a transformer in the \textit{draft stage}. The layouts are further refined by a panoptic refinement network in the \textit{refining stage}. 
}
\label{fig:pipeline}
\end{figure*}


Given a bubble diagram input from the user which encodes the number of rooms, room types, as well as their connections through doors, our proposed framework generates vectorized floorplans in two stages. In the draft stage, we employ a GCN for the layout coordinate embeddings to encode the connectivity relationship between rooms and their connecting doors. The GCN-processed embeddings as well as the category embeddings are then passed into an autoregressive transformer network to generate room layout elements. In the generated floorplan, each room or door is represented by a set of bounding box coordinates. In the refining stage, we apply another transformer network to integrate the global information from the entire floorplan sequence to refine the layouts of the initial floorplan generated in the draft stage. An overview of the proposed framework can be found in Fig.~\ref{fig:pipeline}. More details are introduced in the remaining part of this section and implementation details can be found in the supplementary materials.

\subsection{Draft stage with GCN-constrained autoregressive transformer}\label{GCN-auto}
\subsubsection{Layout representation}
Previous layout generation works~\cite{gupta2021layouttransformer,kong2021blt} represented each object with $5$ attributes $(c, x, y, w, h)$, where $c\in C$ is the category, $(x, y)$ is the center location and $(w, h)$ is the object size. We adopt a similar setting but with the exception of $c$ since in our work, $c$ is given as part of the input conditions encoded in the input bubble diagram therefore there is no need to generate $c$.
$c$ can be directly obtained for each room from the input, where doors are treated as special cases of rooms. 
As illustrated in Fig.~\ref{fig:pipeline}, we use the geometric representation $(x^{L}, y^{T}, x^{R}, y^{B})$ for an object, where $(x^{L}, y^{T})$ is the top-left corner and $(x^{R}, y^{B})$ is the bottom-right corner. We then discretize the continuous $x$- and $y$-coordinates with 8-bit uniform quantization~(from $0$ to $255$)~\cite{gupta2021layouttransformer,nash2020polygen} for the ease of representation. 
Following the convention of autoregressive generation~\cite{gupta2021layouttransformer}, we include an additional starting token and an ending token for the layout sequence. We flatten the whole layout with a sequence of 1-D vectors as:
\begin{equation}\label{eqn:seq}
S = [\langle BoS \rangle, \boldsymbol{r}_1, \boldsymbol{r}_2, ..., \boldsymbol{r}_N, \langle EoS \rangle],
\end{equation}
where $N$ is the total number of rooms and doors, $\langle BoS \rangle$ and $\langle EoS \rangle$ are the starting and ending token of the sequence, respectively. Room or door vector is represented by $\boldsymbol{r}_i = [x^{L}_i, y^{T}_i, x^{R}_i, y^{B}_i]$. Geometric coordinate embedding $X_{g}$ is learned through a learnable embedding layer which projects the original quantized coordinates to an embedding space. 

Also, the room categories corresponding to the objects in the layout sequence are encoded in a separate room category sequence. Following~\cite{gupta2021layouttransformer}, we encode the categories $c$ as discrete values in the range of~$[256, 256+N_{c}]$, where $N_{c}$ is the number of room categories. Then, a category embedding $X_{c}$ is also learned through a learnable embedding layer sharing weights with the geometric embedding layer to convert quantized category values to the embedding space.
\subsubsection{Encoding connectivity graph with GCN}
\begin{figure*}[!tbp]
\centering
\includegraphics[width=0.90\linewidth]{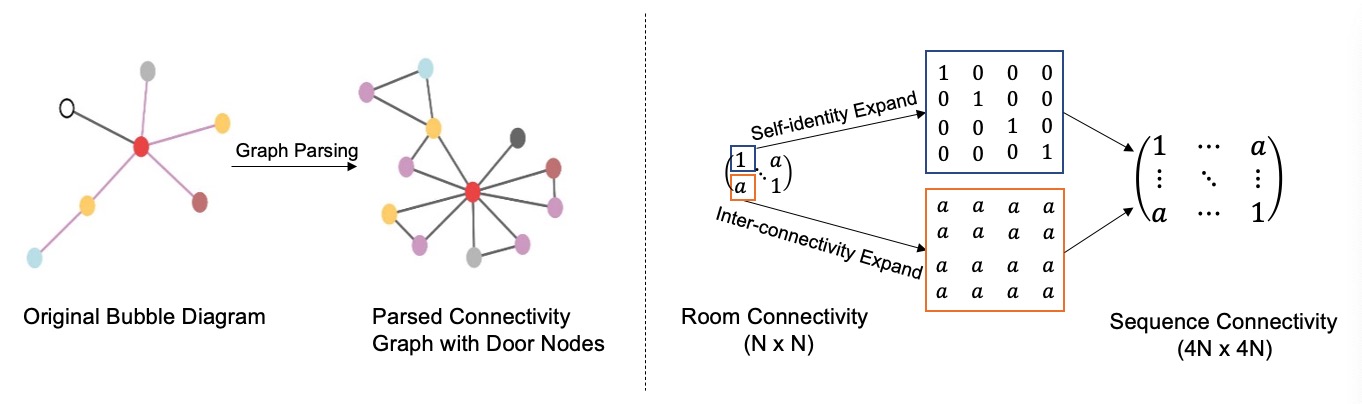}
\caption{{\bf Left}: the process of parsing an original bubble diagram to a connectivity graph adapted for GCN. {\bf Right}: the process of transforming the room connectivity matrix $A$ into the sequence connectivity matrix $A_{S}$.}
\label{fig:graph_parsing}
\end{figure*}

To encode the connectivity relationships, we build a connectivity graph $A_{S}$ which considers the connectivity between rooms via doors from the input bubble diagram. Throughout the paper, we treat all doors as special cases of rooms, as the nodes of the input graph. As illustrated in the left side of Fig.~\ref{fig:graph_parsing}, two nodes are parsed as connected by an edge if there are two rooms connected via a shared door or if a room and a door are connected. Inspired by the success of GCN~\cite{kipf2016semi} in various tasks, we propose to encode the connectivity graph from the bubble diagram with a GCN. The GCN takes the input geometric embedding $X_{g}$ as well as a pre-computed sequence connectivity matrix $A_{S}$ as input. Let $\tilde{A}_{S} = A_{S} + I$, where $I$ is the identity matrix indicating that each room is connected to itself, we adopt a learnable layer $W$ to get the updated feature representation $X_{g}^{G}$  using the row-normalized connectivity matrix $\hat{A}_{S}$ as:
\begin{equation}\label{eqn:GCN}
X_{g}^{G} = f(X_{g}, A_{S}) = \hat{A}_{S}X_{g}W, ~~ \hat{A}_{S} = \tilde{D}^{-\frac{1}{2}}\tilde{A}_{S}\tilde{D}^{-\frac{1}{2}}, ~~ \tilde{D}_{ii} = \sum\nolimits_j \tilde{A}_{S_{ij}}.
\end{equation}

Now we present how to generate the sequence connectivity matrix $A_{S}$ as in the right side of Fig.~\ref{fig:graph_parsing}. For efficient processing by GCN, we adapt the original room connectivity matrix $A\in{N}\times{N}$, where $N$ is the room number, to a sequence connectivity matrix $A_{S}$ for the vectorized floorplan sequence. We then flatten the quantified $N\times4$ bounding boxes to a sequence of 1-D vectors with length $4N+2$, including special tokens $\langle BoS \rangle$ and $\langle EoS \rangle$. For each element $\boldsymbol{r}_{i}$, we get its connectivity value with another element $\boldsymbol{r}_{j}$ by querying $A_{i^{'}j^{'}}$ as $A_{S_{ij}} = A_{i^{'}j^{'}}$, 
where $i^{'}$ and $j^{'}$ are the room indexes of $\boldsymbol{r}_{i}$ and $\boldsymbol{r}_{j}$. For $\langle BoS \rangle$ and $\langle EoS \rangle$, we fox their connectivities to other elements as $0$. $A_{S}$ is used as the input of GCN. Note that during an intermediate step $t$ of the autoregressive generation process, only the generated sequence embeddings $X_{g_{1:t}}$ and a partial graph $A_{S_{1:t}}$ are used as the input of GCN. Thus, the initial floorplan generated in the draft stage may not fully utilize the information of the entire sequence and neglect certain connectivities between rooms. We discuss how to resolve this issue in Sec.~\ref{sec:global-refine}.

\subsubsection{Autoregressive generation with self-attention transformer}
Following previous autoregressive generation works~\cite{gupta2021layouttransformer,nash2020polygen,para2021generative}, we generate the floorplan sequence through an iterative process. The joint distribution of autoregressive generation can be represented by the chain rule:
\begin{equation}
p(s_{1:4N+2}) = \prod_{i=1}^{4N+2}p(s_{i}|s_{1:i-1}).
\end{equation}
A transformer network is applied for autoregressive generation, with a teacher-forcing strategy during training. For the input, we have obtained the GCN-processed geometric embedding $X_{g}^{G}$ and the category embedding $X_{c}$ for each element. To help encode the geometric position of each room inside the floorplan sequence, we also build another embedding layer to learn the positional embeddings $X_{p}$~\cite{vaswani2017attention} for each room element. The input of the transformer is the summation of the three types of embeddings: $X = X_{g}^{G} + X_{c} + X_{p}$.
At the test time, student-forcing is used, and only the starting token is used as input to the transformer network. For the transformer architecture, we apply multi-head self-attention as in LayoutTransformer~\cite{gupta2021layouttransformer} and use GPT-1~\cite{radford2018improving} as our backbone transformer network. Note that our model is flexible and also compatible with other transformer models.

\subsection{Refining stage with panoptic refinement network}\label{sec:global-refine}
A major limitation of transformer-based autoregressive generation is that, for each generation step, only previous states can be taken as input. Thus, the model cannot leverage information from the future states, and may neglect the connectivity between certain elements. For example, in vectorized floorplan generation, it is challenging to properly place a room or door even for human experts, if only limited context information is provided. However, if the initial states for all rooms are given, a transformer model can access the panoptic features and has a better sense of their relative positions with corresponding categories. This motivates us to design a panoptic refining network to improve the floorplan generated in the draft stage by encouraging correct global connectivities.

In the refining stage, we introduce a novel panoptic refinement network (PRN) to refine the initial generations from the autoregressive transformer. The network is also implemented as a GCN and a transformer network with multi-head self-attention modules. Although the network architectures of the two stages are similar, their learning targets are fundamentally different. In the draft stage, our model aims to predict one element at a time given all previously generated rooms; while in the multi-round refining stage, the PRN aims to learn the refined elements based on the entire sequence. 

During training, we use greedy decoding to convert the learned probabilities into the coordinate sequence with discrete values. Note that we do not include $\langle BoS \rangle$ and $\langle EoS \rangle$ in the input of PRN. For the input of the refinement network, since the category embeddings and positional embeddings remain unchanged, we directly get them from the input of the draft stage. Only the geometric embeddings will be updated, which is denoted as $X_{gr}$. The refinement GCN processes $X_{gr}$ and obtain the GCN-updated embeddings $X_{{gr}}^{G}$. Then the input embeddings $X_{r}$ for PRN transformer can be represented in a similar manner as the draft stage: $X_r = X_{{gr}}^{G} + X_{c} + X_{p}$.

Finally, the embedding sequence is passed into the PRN transformer to get the refined floorplan sequences. The end-to-end refinement process of PRN can be performed iteratively by taking the refined output from iteration $i$ as the input of the next iteration $i+1$. More analysis is given in Sec.~\ref{sec:ablation}.

\subsection{Training strategies and losses}\label{loss}
\subsubsection{Hybrid sorting strategy}
As shown in previous works~\cite{nash2020polygen}, ordering is important for training autoregressive models. Layout generation methods~\cite{gupta2021layouttransformer,nash2020polygen} often sorted the elements by the relative spatial positions. However, this strategy used in unconditional generation cannot be directly applied to category-conditioned generation. Since conditional categories are provided as constraints, a category needs to be strictly associated with a generated element. Thus, we propose a hybrid sorting strategy for a more effective conditional autoregressive generation in both training and testing. To leverage categorical spatial information, we first compute the average spatial position for each room category throughout the whole training set, then sort room categories by their average positions and place two types of doors~(\ie{}, interior and front doors) at the end.

\subsubsection{Reconstruction loss}
We apply the standard cross-entropy loss on the generations from both stages to learn a categorical distribution for the quantized coordinates. The loss is averaged by the sequence length $T$ and the refinement iterations $N_{r}$, which can be represented as:
\begin{equation}\label{reconstruction}
L_{recon} = \frac{1}{T}\sum_{t=1}^{T}L_{ar} + \frac{1}{TN_{r}}\sum_{t=1}^{T}\sum_{i=1}^{N_{r}}L_{ref},
\end{equation}
where $L_{ar}$ denotes for the loss from autoregressive model and $L_{ref}$ denotes for the loss from panoptic refinement model.

\subsubsection{Geometric loss} Since the reconstruction loss focuses more on single elements, the connectivity between rooms is not explicitly measured by the reconstruction loss. To further encourage correct room layout and connectivity, especially for the refining stage, we introduce a global geometric constraint by calculating the intersection-over-union (IoU) values between the generated and real layouts. To calculate the IoU values, we need to first transform the predicted probabilities on the quantized coordinates to continuous values. However, direct conversion is not differentiable and thus cannot be trained via back-propagation. To enable end-to-end training, we borrow ideas from the differentiable soft-argmax strategy originally proposed in the stereo matching task~\cite{kendall2017end}. The $i$th element's prediction $x_{i}$ is then represented as $x_{i} = \sum_{j=1}^{N}x_{j}\cdot \sigma(x_{j}),$
where $\sigma({x_{j}})$ is the probability of quantized coordinate ${x_{j}}$ implemented with the softmax function. Now we can compute a geometric IoU matrix $\mathcal{G}_{gen}$ for the generated sequence and build a geometric loss $L_{geo}$ measuring the $L_1$ distance between $\mathcal{G}_{gen}$ and its groundtruth geometric IoU matrix $\mathcal{G}_{gt}$:
\begin{equation}
L_{geo} = \frac{1}{N_{l}}\sum_{i_{l}=1}^{N_{l}}~||\mathcal{G}_{gen}^{i_{l}} - \mathcal{G}_{gt}^{i_{l}}||_{1}\;,
\end{equation}
where $i_{l}$ and $N_{l}$ are the index and the total number of pair-wise objects in a floorplan, respectively. $L_{geo}$ is applied in both draft stage and refining stage. Our final loss objective is the combination of two losses without scaling factors.
\section{Experiments}




\subsection{Dataset and pre-processing}\label{pre-process}
Similar to~\cite{nauata2021house,para2021generative}, we conducted experiments on RPLAN dataset~\cite{wu2019data} which provides vector-graphic floorplans from real-world residential buildings. We followed the pre-processing pipeline used in HouseGAN++ with a modification for bounding-box~(bbox) vectorized floorplan generation. When we transformed the non-standard polygons into rectangular bounding boxes, we found some cases where frontal doors or interior doors were incorrectly kept inside a room (see Fig.~1 in supplementary materials as an example). To filter out such noisy cases, we computed the overlapping between doors and rooms, then discarded a sample if its frontal doors overlapped with any rooms or any of its interior doors overlapped with any non-living rooms over a threshold $\tau = 50\%$.

\subsection{Baselines and evaluation metrics}
We compared our method with two state-of-the-art floorplan generation methods using the same input setting~\cite{nauata2020house,nauata2021house} as ours. We also compared with LayoutTransformer~\cite{gupta2021layouttransformer} which we adopted as the backbone generation network. For fair comparisons, we retrained all baseline methods with their official implementations and applied the same data pre-processing as described in Sec.~\ref{pre-process}.

Following~\cite{nauata2020house,nauata2021house} which also took bubble diagram as input constraints, we evaluated our method in terms of realism, diversity, and compatibility. For realism, we invited 10 respondents with architecture design expertise~(professors and graduate students from the architecture department) and 7 amateurs to give rankings on 100 generation examples based on visual quality and functionality. Our method and two other competing methods~\cite{nauata2021house,gupta2021layouttransformer} each generated 100 floorplans and respondents ranked the methods after reviewing each set of 3 generated floorplans that correspond to the same input bubble diagram. 
A method obtained a score $+1$ for being ranked as $1st$, $0$ for $2nd$, and $-1$ for $3rd$. Diversity was computed as the FID score~\cite{heusel2017gans} between the generated floorplans from a method and the real floorplans, to assess the diversity of the generated floorplans given the same input. The compatibility~\cite{nauata2021house} validates the connectivity correctness of a generated floorplan by computing the graph edit distance (GED)~\cite{abu2015exact} between the input graph with the reconstructed graph from the generated floorplan. For a generated plan to achieve a high compatibility score, a door must be correctly placed between two rooms (or between a room and the outside area). We generated $1,000$ testing samples on each data split and ran each model for $5$ rounds to get the mean and standard deviation of FID and GED scores.

\begin{figure*}[!tbp]
\centering
\includegraphics[width=0.95\linewidth]{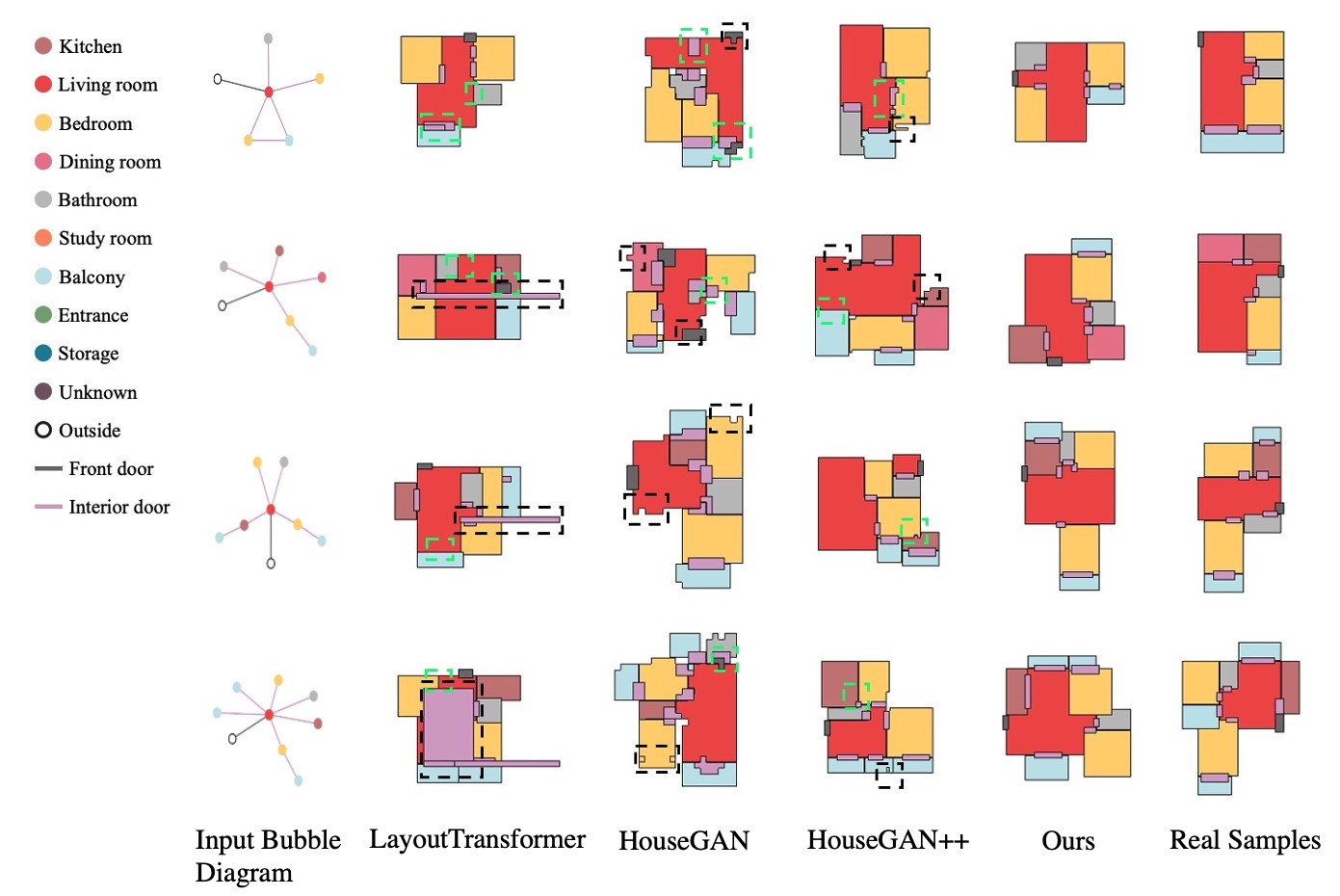}
\caption{Sampled floorplan generations from different methods, including real floorplans and their corresponding input bubble diagrams. We use the same color legend as in~\cite{nauata2021house}. Areas with salient connectivity issues and visual artifacts are highlighted with {\color{green}green} and \textbf{black} dashed boxes, respectively.}
\label{fig:qualitative}
\end{figure*}

For a comprehensive comparison, we evaluated our methods under both the \textit{separate training} and \textit{mixed training} settings. Under the separate training setting~\cite{nauata2021house}, floorplans were split into training and testing sets based on different numbers of rooms. Training was done using any number of rooms except the number of rooms reserved for testing. Comparisons using the separate training are provided in Table~\ref{tab:sota}. In addition to the separate training, we also reported metrics calculated using the \textit{mixed training}. Specifically, we mixed floorplans with different numbers of rooms together and split them into training and testing without overlapping in input bubble diagrams. We believe this is a more general setting since floorplans with different number of rooms should all be available for training. Comparisons using mixed training are illustrated in Table~\ref{tab:user_study}.

\begin{table*}[tbp]
\centering
\caption{Quantitative comparison between methods using the mixed training. Diversity and compatibility metrics are reported as ``mean$/$std.''. }
\label{tab:user_study}
\resizebox{0.90\linewidth}{!}{
\begin{tabular}{l|c|c|c|c}
\hline
{Model} & {Visual Realism~$\uparrow$} & {Functional Realism~$\uparrow$} & {Diversity~$\downarrow$} & {Compatibility~$\downarrow$}\\
\hline 
LayoutTransformer~\cite{gupta2021layouttransformer} & -0.58  & -0.57 & \textbf{12.3/0.2}  & 4.6/0.0\\
HouseGAN++~\cite{nauata2021house} & 0.12 & 0.09 & 15.7/0.2  & 3.5/0.0\\
Ours   & \textbf{0.46} & \textbf{0.48} & 13.8/0.4  & \textbf{2.8/0.1}\\
\hline
\end{tabular}
}
\end{table*}

\subsection{Result Analysis}

\begin{table*}[tbp]
\centering
\caption{Quantitative comparison between different methods using the separate training. Integers refer to train on floorplans with other number of rooms and test with the designated number of room.}
\label{tab:sota}
\resizebox{0.90\linewidth}{!}{
\setlength{\tabcolsep}{4pt}{
\begin{tabular}{l|cccc|cccc}
\hline
 \multirow{2}{*}{Model} & \multicolumn{4}{c|}{Diversity~$\downarrow$} & \multicolumn{4}{c}{Compatibility~$\downarrow$}\\
& $5$ & $6$ & $7$ & $8$ & $5$ & $6$ & $7$ & $8$ \\
\hline 
LayoutTransformer~\cite{gupta2021layouttransformer} & 21.5/0.3 & 27.8/0.5 & 25.6/0.6 & 28.4/0.8 & 2.4/0.1 & 4.0/0.1 & 5.6/0.1 & 6.4/0.1\\
HouseGAN~\cite{nauata2020house} & 57.6/1.3 & 70.1/1.2 & 61.7/0.9 & 57.7/0.6 & 3.5/1.6 & 4.3/1.7 & 5.2/1.9 & 6.1/1.9\\
HouseGAN++~\cite{nauata2021house} & 24.1/0.6 & 17.7/0.7 & 15.0/0.7 & 22.2/0.4 & 1.6/0.0 & 2.5/0.1 & \textbf{2.8/0.0} & \textbf{3.8/0.0}\\
Ours   & \textbf{15.4/0.5} & \textbf{16.4/0.5} & \textbf{14.6/0.4} & \textbf{14.1/0.4} & \textbf{1.3/0.0} & \textbf{1.9/0.0} & 2.9/0.1 & 4.6/0.0 \\
\hline
\end{tabular}
}
}
\end{table*}

\begin{figure*}[tbp]
\centering
\includegraphics[width=0.9\linewidth]{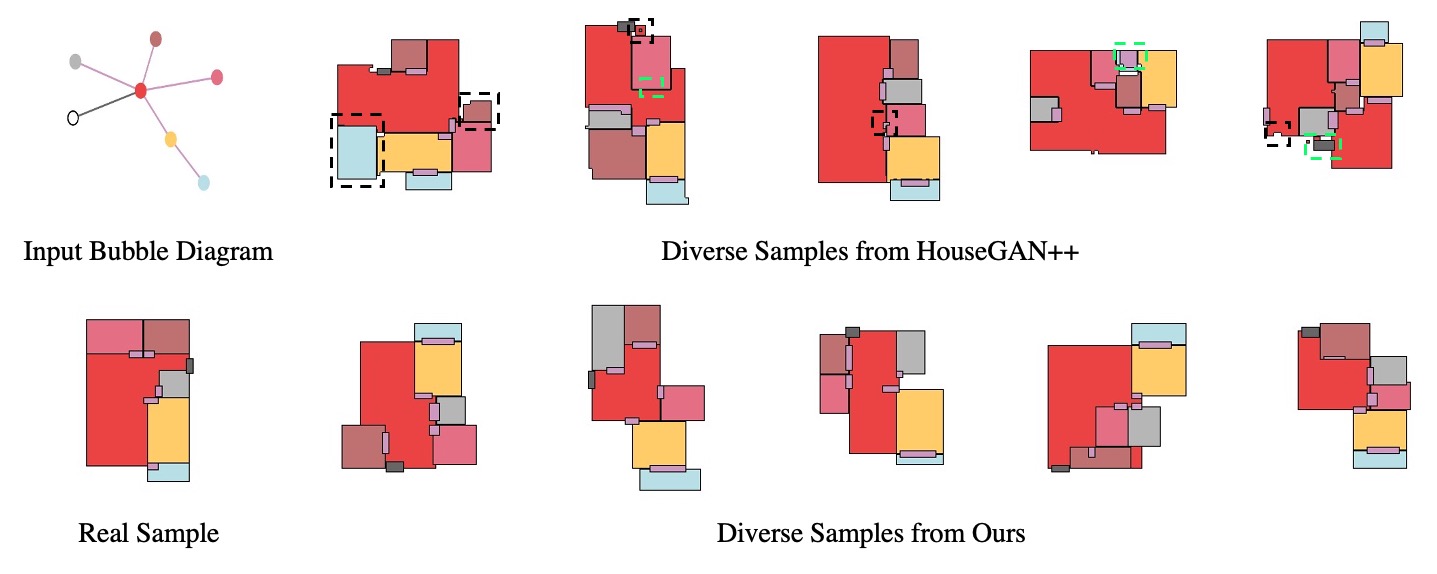}
\caption{Diversity comparison between HouseGAN++ and ours. Salient connectivity issues and visual artifacts are highlighted with {\color{green}green} and \textbf{black} boxes.}
\label{fig:diversity}
\end{figure*}

Table~\ref{tab:sota} shows the quantitative comparisons among different methods under the separate training setting. Our method outperforms all the previous methods on diversity by a large margin. For compatibility, our model is significantly better than existing approaches on the split containing $5$ or $6$ rooms, and is comparable with HouseGAN++~\cite{nauata2021house} on the $7$ room split. Under our proposed mixed training setting in Table~\ref{tab:user_study}, which we believe is more suitable for realistic scenarios, our method clearly outperforms HouseGAN++ on all metrics. Note that evaluation of visual and functional realism by ten architecture experts and seven amateurs was conducted using floorplans generated in the mixed training setting, and the result on realism comparison is reported in Table~\ref{tab:user_study}. One can see that, for realism, our method outperforms~\cite{gupta2021layouttransformer,nauata2021house} on both visual quality and functionality by a large margin.
Qualitatively, from the samples shown in  Fig.~\ref{fig:qualitative},  one can see that LayoutTransformer~\cite{gupta2021layouttransformer} often fails to generate doors correctly, especially when the input bubble diagrams are complex. The rooms generated by HouseGAN~\cite{nauata2020house} have salient artifacts due to the limitations of rasterized representation. There are fewer artifacts in floorplans generated by HouseGAN++~\cite{nauata2021house}, but non-straight room boundaries still exist. Another non-negligible drawback of rasterized representation is that several disconnected regions could be generated for one specific room~(\eg{}, the living room of the $3rd$ case by HouseGAN++, Fig.~\ref{fig:qualitative}).
In comparison, our method only generates objects as connected components due to the advantage of vectorized representation. Furthermore, as shown in Fig.~\ref{fig:diversity}, our model better keeps the fidelity and connectivity while generating diverse samples than the state-of-the-art HouseGAN++~\cite{nauata2021house}.

\subsection{Ablation study}\label{sec:ablation}

\subsubsection{Proposed components.}
\begin{table*}[tbp]
\caption{Ablation study of our proposed method. Base refers to the backbone GPT-1~\cite{radford2018improving} transformer network. HS, PR, GL represent our proposed hybrid sorting, panoptic refinement, geometric loss, respectively.}
\centering
\begin{minipage}[t]{0.49\linewidth}
\subcaption{Our proposed components.}
\label{tab:ablation}
\centering
\resizebox{0.95\linewidth}{!}{
\begin{tabular}{ccccc|c|c}
\hline
\multicolumn{5}{c|}{Components} & {Diversity~($\downarrow$)} & {Compatibility~($\downarrow$)}\\
Base & HS & GCN & PR & GL & 6 & 6\\
\hline 
\checkmark & & & & & 27.8/0.5 & 4.0/0.1\\
\checkmark & \checkmark & & & & 29.8/0.6 & 3.6/0.1 \\
\checkmark & \checkmark & \checkmark & & & 14.5/0.2 & 3.7/0.0 \\
\checkmark & \checkmark & & \checkmark & & \textbf{14.0/0.3} & 3.4/0.1 \\
\checkmark & \checkmark & \checkmark & \checkmark & & 18.3/0.2 & 2.1/0.1\\
\checkmark & \checkmark & \checkmark & \checkmark & \checkmark & 15.8/0.5 & \textbf{1.9/0.0}\\
\hline
\end{tabular}}
\end{minipage}
\begin{minipage}[t]{0.49\linewidth}
\subcaption{Refinement iterations of PRN.}
\label{tab:refine_iters}
\centering
\resizebox{0.95\linewidth}{!}{
\begin{tabular}{l|c|c}
\hline
\multirow{2}{*}{{\# of iteration}}& \multicolumn{1}{c|}{Diversity~($\downarrow$)} & \multicolumn{1}{c}{Compatibility~($\downarrow$)}\\
 & $6$ & $6$\\
\hline
1-iteration & 15.1/0.8 & 2.7/0.0\\
3-iterations & \textbf{14.4/0.3} & 2.2/0.0\\
5-iterations & 15.8/0.5 & \textbf{1.9/0.0}\\
7-iterations & 16.9/0.4 & 2.1/0.1\\
\hline
\end{tabular}
}
\end{minipage}
\end{table*}

We conducted an ablation study on each of our proposed components, including the hybrid sorting (HS) strategy, the use of GCN to encode connectivity graph, the proposed panoptic refinement (PR) network, and the geometric loss (GL). Specifically, we conducted ablation experiments on the 6 room generation split. The base model corresponds to the result of LayoutTransformer~\cite{gupta2021layouttransformer}, which we used as the unconditional backbone network. As shown in Table~\ref{tab:ablation}, applying the hybrid sorting strategy improved the compatibility compared with the base model. Adding GCN or refinement model independently upon the autoregressive model significantly improved the diversity, but had little benefits for compatibility. A possible reason is that, passing a partial graph containing previously generated elements to GCN can smooth the feature for past elements, but is not beneficial to the generation for future elements. Applying the two modules simultaneously greatly improved compatibility since a complete connectivity graph drives GCN to capture the connectivity relationship of the whole sequence, and the refinement transformer can further exploit global contexts. With the proposed geometric loss on the generated vectors, both diversity and compatibility got further improved as the correct layouts and room connectivities were encouraged during the training.

\subsubsection{Refinement iterations}

\begin{figure*}[!t]
\centering
\includegraphics[width=0.8\linewidth]{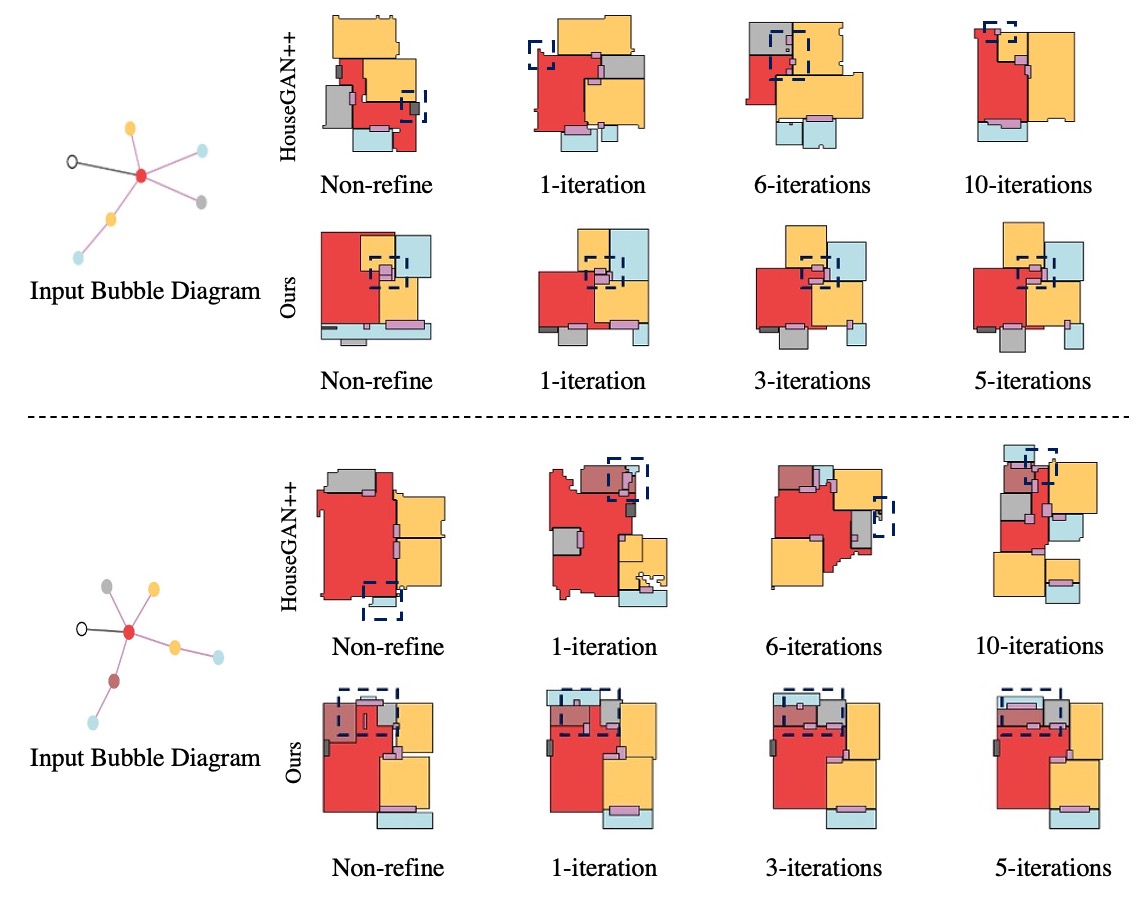}
\caption{Floorplan refinement between ours and~\cite{nauata2021house}. Salient differences across different iterations are highlighted with \textbf{black} dashed boxes.}
\label{fig:refine_iters}
\end{figure*}

We also demonstrate the effect of applying different numbers of refinement iterations. As shown in Table~\ref{tab:refine_iters}, there were no significant differences in diversity among various iterations. For compatibility, the best result was achieved when we refined the initial generated sequence for $5$ iterations. We have tried to refine more iterations, such as $N_r = 7$, but the experiments showed no further improvement, which may indicate that PRN has adequately refined the draft floorplan. We also show some qualitative visualizations in Fig.~\ref{fig:refine_iters} to help examine the effect of our multi-round refinements.

\section{Discussion}
HouseGAN++~\cite{nauata2021house} also proposed to refine floorplans iteratively using heuristics and optimizations. Our proposed PRN refinement network is very different, however, since it is an end-to-end model which refines the draft floorplan  while keeping correct layouts. As illustrated in Fig.~\ref{fig:refine_iters},  applying one iteration of PRN refinement (\ie{}, 1-iteration) significantly improved the generation quality compared with no refinement (\ie{}, `Non-refine'). Refining more iterations consistently made details more accurate, such as reducing overlaps, making neighboring objects better aligned, and placing doors more properly. Compared with the floorplans after multiple refinement iterations by HouseGAN++, the floorplans refined by our PRN were more consistent in layout and inherited correct room connectivities acquired in previous iterations of refinement. Meanwhile, refinements by HouseGAN++ introduced more drastic changes across different iterations, and occasionally regions with desirable properties from previous iterations got lost and disappeared in the later iterations.

Although we have achieved superior performance under most settings, our method still has limitations. First, our model represents rooms as rectangular boxes, which may not always be optimal in practice. We try to mitigate the issue by removing overlapping parts between the living room and other rooms, making the generated floorplan more general. An end-to-end generation with vectorized multi-edge polygons representing rooms could be a future direction. Second, we did not achieve better compatibility performance than HouseGAN++ on the split of 7 or 8 rooms, indicating that our method can be further improved for modeling longer floorplan sequences. More advanced GNNs and transformers may perform better in modeling more complex graphs and longer sequences.

During the user study, besides rankings of methods which we used to calculate realism scores, we also collected feedback in the form of free-form comments from architects. Most of the architects agreed that our method has made substantial progress in automated floorplan generation, and can potentially change the way in which initial floorplan designs are created and visualized. They especially liked that methods like ours can easily generate multiple candidate initial designs. However, they also mentioned that certain domain knowledge should be embedded in the generation process, for example, circulation spaces must be smaller, kitchen and dining must be connected, service blocks like washroom should be at the edge of the building. We plan to explore how to incorporate such domain knowledge into the generation process in our future work.

\section{Conclusion}
In this paper, we propose a novel two-stage vectorized floorplan generation framework. Floorplans can be generated and refined in an end-to-end fashion given an input bubble diagram encoding room number, type and connectivity. Compared to previous state-of-the-art floorplan generation methods which generate rasterized floorplan images, our proposed vectorized floorplan generation method generates more realistic and usable floorplan designs. Endorsed by the architect experts in our user study, we believe our method has great potential in computer-aided floorplan design. 

\medskip
\noindent{\bf Acknowledgements.} This work is supported in part by NSF Award \#1815491. We appreciate the help from professors and graduate students from College of Arts and Architecture at Penn State with the user study. We also would like to thank Enyan Dai for meaningful discussions on GNN.

\clearpage
%
%
\bibliographystyle{splncs04}
\bibliography{2645}
\end{document}


\pagestyle{headings}
\mainmatter
\def\ECCVSubNumber{2645}  

\title{Supplementary Materials for: \\ End-to-end Graph-constrained Vectorized Floorplan Generation with Panoptic Refinement} 


\titlerunning{Supp: Vectorized
Floorplan Generation with Panoptic Refinement}
%
\author{}
%
\authorrunning{J. Liu et al.}
%
\institute{}
\maketitle

\section{Implementation details}

We implemented our proposed method using PyTorch~\cite{paszke2019pytorch} and ran all experiments on an NVIDIA RTX 3090 GPU. We set up the codebase based on the official implementation of LayoutTransformer~\cite{gupta2021layouttransformer}. The AdamW~\cite{loshchilov2018decoupled} optimizer was applied with a constant learning rate $3\times10^{-4}$. We trained our model end-to-end for 20 epochs with batch size $128$. We applied student forcing to generate the draft floorplan sequence from scratch during testing, starting from the $\langle BoS \rangle$ token. Specifically, we used greedy decoding with top-$k$ probabilities, then sampled from the probability distribution to obtain discrete coordinates. We set $k=5$ in all experiments. All feature dimensions used in the proposed network were set to be $128$. Based on experimental results, we chose $5$ refinement iterations as the default setting in the refining stage.

\begin{figure*}[!hbp]
\centering
\includegraphics[width=0.8\linewidth]{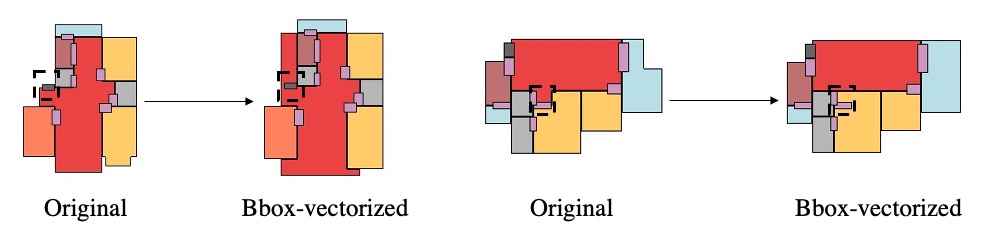}
\caption{Examples of noisy data introduced by bbox vectorization.}
\label{fig:noisy_data}
\end{figure*}

\section{Model size and speed comparison}
As shown in Table~\ref{tab:speed_parameter}, compared with the two other representative works~\cite{gupta2021layouttransformer,nauata2021house}, although our model has more parameters, it achieves a good trade-off among inference time, memory consumption, and generation performance. 
Although transformer models are parameter-heavy, our method is only slightly slower than~\cite{gupta2021layouttransformer} because the proposed panoptic refinement does not add much overhead compared to the autoregressive generation process.

\begin{table}[!tbp]
\caption{Number of parameter and speed comparison among different methods.}
\label{tab:speed_parameter}
\resizebox{1.00\linewidth}{!}{
\begin{tabular}{l|c|c|c}
\hline
{Methods} & {Parameters~/~M} & {Speed~/~FPS $\uparrow$} & {Memory~/~G $\downarrow$} \\
\hline 
LayoutTransformer~\cite{gupta2021layouttransformer} & 1.27  & 5.65 & 1.25\\
HouseGAN++~\cite{nauata2021house} & 0.64 & 2.83 & 1.6\\
Ours   & 2.51 & 4.22 & 1.27\\
\hline
\end{tabular}
}
\end{table}

\section{User interaction and design customization}
\begin{figure}[!htp]
\flushleft
\includegraphics[width=1.0\linewidth]{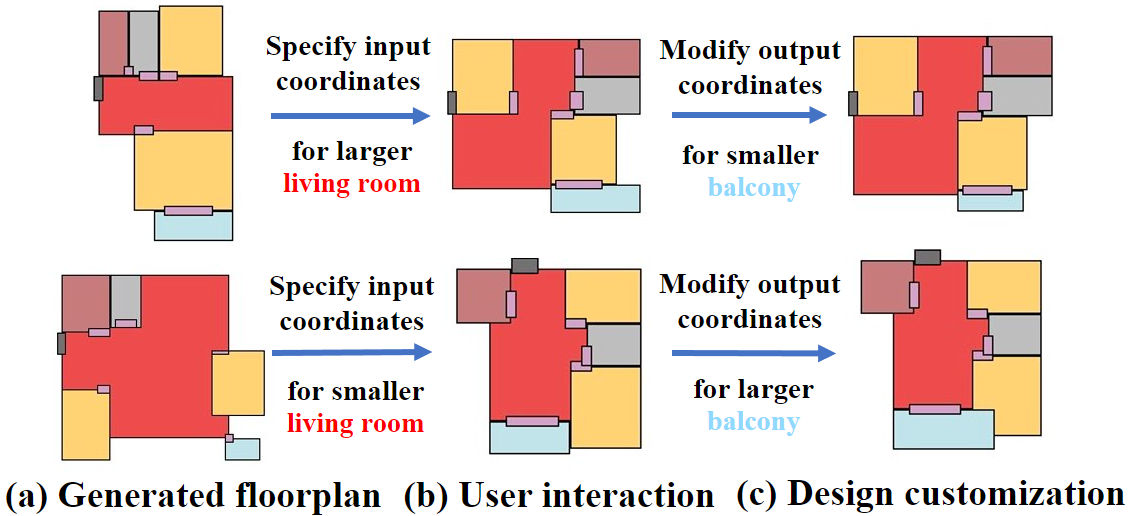}
\caption{Examples on user interaction and design customization.}
\label{fig:customization}
\end{figure}

Our vectorized floorplan generation method allows flexible user interaction and design customization. For user interaction, our method generates multiple outputs with single input for user to choice from, which was greatly appreciated by design professionals during our user study.
Our framework also allows users to easily specify input coordinates for certain rooms to customize size and location (see Fig.~\ref{fig:customization}(b) below).
Further, for design customization, the vectorized representation makes it easy to customize the size or location of rooms by directly modifying generated room coordinates (see Fig.~\ref{fig:customization}(c)).

\clearpage
%
%
\bibliographystyle{splncs04}
\bibliography{2645}